\documentclass[10pt,twocolumn,letterpaper]{article}

\usepackage{iccv}              
\usepackage[pagebackref,breaklinks,colorlinks]{hyperref}
\usepackage{multirow}
\usepackage{parskip}
\usepackage{indentfirst}
\title{Learning Adaptive Pseudo-Label Selection for Semi-Supervised 3D Object Detection}

\author{
Taehun Kong$^{1,2}$ \, and \, Tae-Kyun Kim$^{1}$\\[1mm]
$^1$School of Computing, KAIST\qquad  $^2$AI Lab, LG Electronics\\
}

\begin{document}
\maketitle
\begin{abstract}
Semi-supervised 3D object detection (SS3DOD) aims to reduce costly 3D annotations utilizing unlabeled data. Recent studies adopt pseudo-label-based teacher-student frameworks and demonstrate impressive performance. The main challenge of these frameworks is in selecting high-quality pseudo-labels from the teacher’s predictions. Most previous methods, however, select pseudo-labels by comparing confidence scores over thresholds manually set. The latest works tackle the challenge either by dynamic thresholding or refining the quality of pseudo-labels. Such methods still overlook contextual information e.g., object distances, classes, and learning states, and inadequately assess the pseudo-label quality using partial information available from the networks. In this work, we propose a novel SS3DOD framework featuring a learnable pseudo-labeling module designed to automatically and adaptively select high-quality pseudo-labels. Our approach introduces two networks at the teacher output level. These networks reliably assess the quality of pseudo-labels by the score fusion and determine context-adaptive thresholds, which are supervised by the alignment of pseudo-labels over GT bounding boxes. Additionally, we introduce a soft supervision strategy that can learn robustly under pseudo-label noise. This helps the student network prioritize cleaner labels over noisy ones in semi-supervised learning. Extensive experiments on the KITTI and Waymo datasets demonstrate the effectiveness of our method. The proposed method selects high-precision pseudo-labels while maintaining a wider coverage of contexts and a higher recall rate, significantly improving relevant SS3DOD methods.
\vspace{-8pt}
\end{abstract}

\section{Introduction}
\label{sec:intro}
3D object detection in LiDAR point clouds has become an essential task for scene understanding in fields such as autonomous driving, robotics, and AR/VR.
Although advanced 3D object detection methods have been developed, they still require a substantial amount of 3D labels. Moreover, high-quality 3D labeling demands precise bounding box coordinates and careful comparison with corresponding 2D images for object class labels. Such labor-intensive 3D labeling processes result in an imbalance, as significantly more data remains unlabeled compared to labeled data. In response to this, semi-supervised learning (SSL) serves as an effective solution, enabling the utilization of unlabeled data to improve performance.

\begin{figure}
    \subfloat[Previous SS3DOD framework]{{\includegraphics[width=0.24\textwidth ]{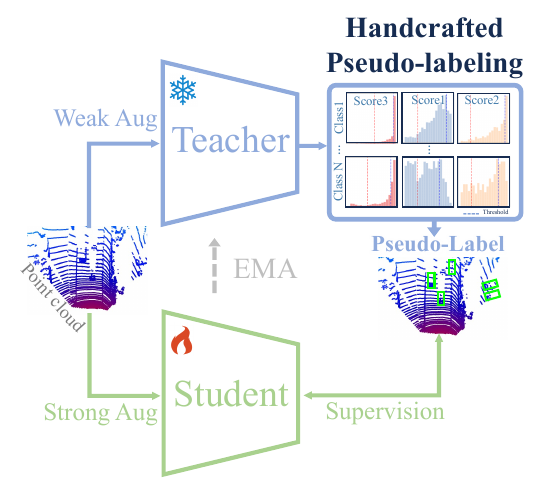} }}%
    \subfloat[Our Framework]{{\includegraphics[width=0.22\textwidth ]{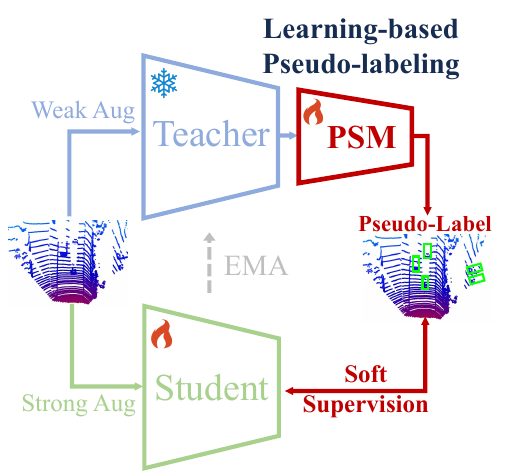} }}%
    \caption{Overview of the proposed framework compared to the
previous pseudo-labeling method in the semi-supervised framework. (a) illustrates the previous semi-supervised framework, where thresholds are determined manually or handcrafted, and filtering is applied based on those thresholds. (b) Shows the proposed framework, which includes the Pseudo-label Selection Module, which learns to select high-quality pseudo-labels within the SSL framework while ensuring robust training against pseudo-label noise through Soft Supervision.}%
    \label{fig:mainfig}%
\vspace{-8pt}
\end{figure}

\begin{figure*}[t]
\centering
\subfloat[Score dist. by classes]{{\includegraphics[width=0.22\textwidth ]{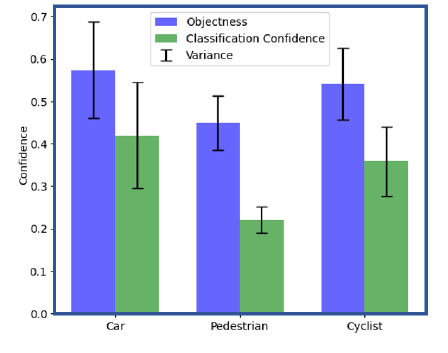} }}%
\subfloat[Score dist. by distance]{{\includegraphics[width=0.22\textwidth ]{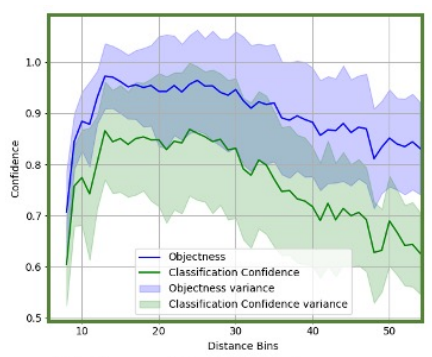}} \label{fig:scoredist_distance}}%
\subfloat[Score dist. by learning states]{{\includegraphics[width=0.22\textwidth ]{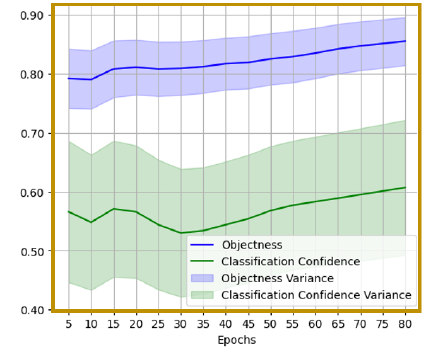} }}%
\subfloat[Comparison of Pseudo-labeling methods]{{\includegraphics[width=0.28\textwidth ]{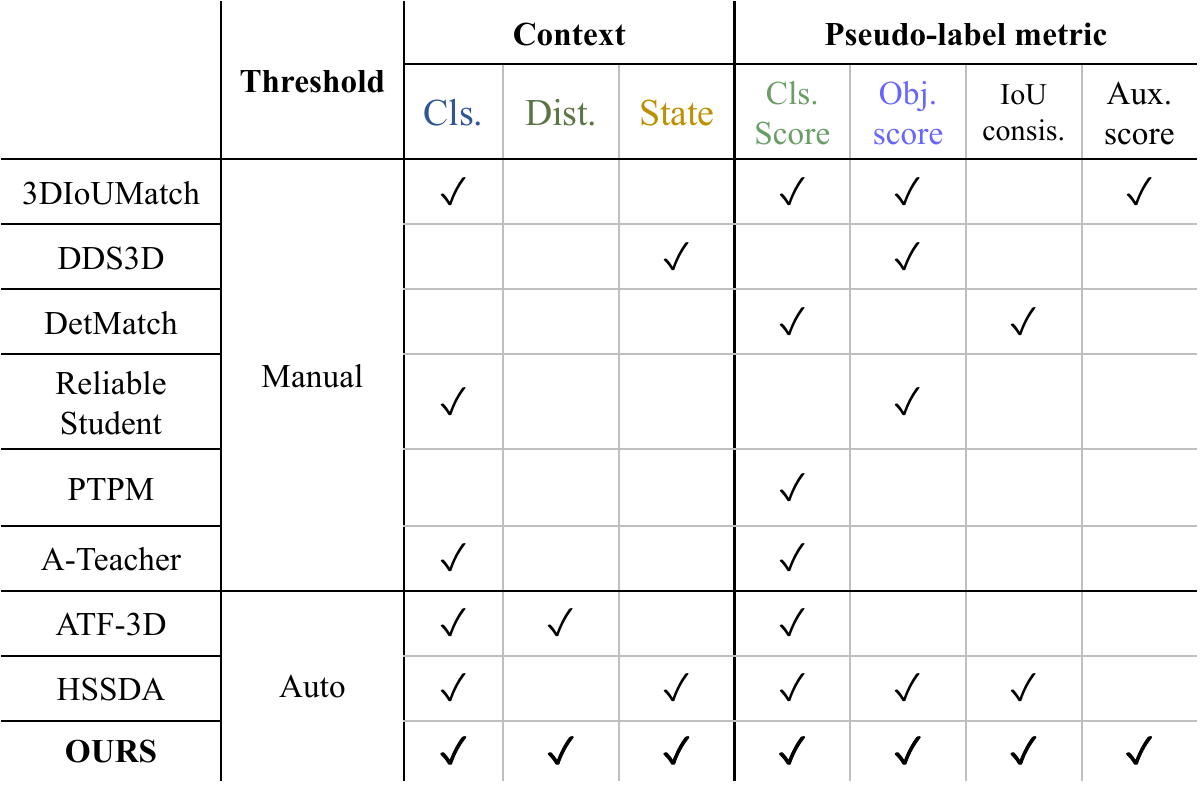} }\label{fig:scoredist_d}}%
\vspace{-5pt}
\caption{(a), (b), and (c) show that classification confidence and objectness have different distributions depending on the context. (b) and (c) illustrate the distributions specifically for foreground objects. (d) compares previous pseudo-labeling methods in three aspects: the approach for determining score thresholds, the contexts considered, and the metrics used for evaluating pseudo-label quality. Auxiliary scores (Aux. score) refer to additional IoU predictions or objectness from different views.}
\label{fig:scoredist}   
\vspace{-8pt}
\end{figure*}

The pseudo-label-based framework has been most widely adopted for Semi-Supervised 3D Object Detection (SS3DOD), effectively leveraging unlabeled data. Variants of SS3DOD methods \cite{3dioumatch,proficient,dds3d,atf3d,detmatch,hssda,ted_hssda,reliable,patchpillar,ateacher} based on this framework have been developed, achieving significant performance gains. Within this framework, the pseudo-labeling plays a critical role in detection performance. 
Previous studies typically select pseudo-labels by thresholding scores predicted by the teacher network (e.g., classification confidence and objectness). For instance, \cite{3dioumatch,dds3d,detmatch,ateacher,reliable,patchpillar} manually set the thresholds to filter pseudo-labels. 
ATF-3D \cite{atf3d} introduced a threshold searching mechanism by object distances and classes.  The state-of-the-art method, HSSDA \cite{hssda}, automatically set dual thresholds. For three object classes, nine clustering steps on three scores determine 18 thresholds for hierarchical supervision. Using these thresholds directly on the teacher’s scores causes suboptimal pseudo-label selection for two reasons. First, predicting the quality of pseudo-labels is challenging. In SS3DOD, the base detector typically yields multiple scores. Each score exhibits a different correlation with GT pseudo-label quality (see \cref{fig:scatter}). This complicates setting a consistent threshold across multiple scores and assessing the reliability of pseudo-labels. As a result, previous methods relied only on partial information rather than utilizing all available indicators for assessing pseudo-label quality (see \cref{fig:scoredist_d}). 
Second, the optimal threshold must account for the context of instances (e.g., object classes, distances, and learning states). \cref{fig:scoredist} illustrates that the score distributions vary across classes, distances, and learning states. The optimal threshold is consequently context-dependent. Rather than a fixed threshold, an effective threshold needs to balance the quality and coverage of pseudo-labels across contexts. Moreover, addressing the dynamic learning states during training requires recalculating the threshold for SSL. 
Finding an optimal threshold that accounts for such contexts is complex, approaches such as \cite{3dioumatch,atf3d,dds3d,hssda} only partially consider these contextual factors (see \cref{fig:scoredist_d})

To address the aforementioned limitations of prior work, we propose a novel learning-based pseudo-label selection method, named Pseudo-label Selection Module (PSM). The PSM leverages limited Ground Truth (GT) information to assess the quality of pseudo-labels and to determine a context-appropriate threshold. The PSM consists of the Pseudo-Label Quality Estimator (PQE) and Context-aware Threshold Estimator (CTE). The PQE encodes the teacher’s various output scores to a single fusion score indicating reliable pseudo-label quality, while the CTE encodes the context to generate adaptive threshold values. During SSL, PSM is trained to dynamically select pseudo-labels considering the context, achieving a wide coverage of pseudo-labels while maintaining a high quality. Additionally, we introduce the Soft Supervision strategy to train robustly against pseudo-label noises. Our method combines the soft GT sampling augmentation and loss re-weighting to counteract pseudo-label noises given the coverage of labels.
To the best of our knowledge, this is the first method to model pseudo-labeling using a neural network. Our contributions are summarized as follows:
\begin{itemize}
    \item We introduce a novel learning-based pseudo-label selection method, Pseudo-label Selection Module (PSM), which better predicts the pseudo-label quality and considers the contexts for pseudo-label selection.
    \item We propose a noise-robust supervision strategy that prevents the student from being biased to pseudo-label noises.
    \item Extensive experiments on the KITTI and Waymo datasets show that the proposed framework significantly improves performance. Notably, in the limited labeled data scenario of KITTI 1\%, we achieved around 20 mAP absolute improvement over the labeled-only 3D baseline.
\end{itemize}
\section{Related Work}
\label{sec:related}

\subsection{3D Object Detection}
\label{ssec:3dod}
3D object detection involves predicting oriented 3D bounding boxes and types of objects from either monocular RGB images \cite{shi2020distance, shi2021geometry, hong2022cross, shi2023multivariate, yan2024monocd} or LiDAR point clouds. 3D object detection from LiDAR point clouds is generally categorized into point-based detectors \cite{pointrcnn,pointgnn,std,3dssd,deephough,pointformer} and grid-based detectors \cite{second, pointpart,voxelrcnn,center,voxeltransformer,pointpillars,pillar} by data representations. Point-based detectors encode spatial information directly from the raw point cloud. In PointRCNN \cite{pointrcnn}, the point-based backbone \cite{pointnet++} hierarchically samples the input point cloud and extracts features. The extracted point-level features are used in a two-stage process to generate and refine 3D proposals. 
In contrast, grid-based detectors convert the sparse point cloud into a grid representation (e.g., voxels and pillars) and then efficiently encode the scene using conventional CNNs. VoxelNet \cite{voxelnet} extracts features from points within voxels using PointNet \cite{pointnet}, and then generates 3D detection proposals using 3D CNNs. SECOND \cite{second} improves speed and efficiency by applying sparse 3D convolutions instead of dense 3D convolutions. Additionally, \cite{pointpillars,pillar} stacks the point cloud into vertical columns (pillars) and applies a simplified PointNet to extract features from each pillar. 
Point-voxel-based detectors combine the strengths of both grid-based and point-based detectors. Studies such as \cite{pvrcnn,pvrcnn++,pyramidrcnn,lidarrcnn,ct3d} utilize voxel- and point-based operations for 3D proposal generation or refinement. In this work, we conducted experiments using the grid-based detector Voxel-RCNN \cite{voxelrcnn} and the point-voxel-based detector PV-RCNN \cite{pvrcnn}.

\subsection{Semi-Supervised Learning (SSL)}
\label{ssec:ssl}
Semi-supervised learning can be broadly divided into two categories: consistency regularization \cite{temporal,meanteacher,virtual,uda} and pseudo-labeling \cite{pseudolabel,labelpropagation,selftraining}. Consistency regularization is a supervision approach that ensures the model’s outputs remain consistent even under different views of the same scene. Mean Teacher \cite{meanteacher} divides the network into a teacher and a student, where the teacher is updated as the student's Exponential Moving Average (EMA). A constraint is applied to make the outputs of both models consistent for data subject to different augmentations. Another key category is the pseudo-labeling, which generates pseudo-labels for unlabeled data to offer more supervision. 
FixMatch \cite{fixmatch} selects high-quality pseudo-labels by manually setting a confidence threshold, while FlexMatch \cite{flexmatch} scales thresholds using the class-specific learning effect. While SemiReward \cite{semireward} introduced two additional networks to measure the pseudo-label reliability via adversarial learning, it still resorts to predefined thresholds for pseudo-label selection. 
By contrast, our method tackles both reliable pseudo-label quality prediction and automatic pseudo-label selection.

\subsection{Semi-Supervised 3D Object Detection}
\label{ssec:ss3dod}
SSL has been extensively studied in 2D object detection \cite{sohn2020simple, xu2021end, zhou2021instant, choi2022semi, wang2023consistent} and is now drawing increasing attention in 3D object detection. SESS \cite{sess} and 3DIoUMatch \cite{3dioumatch} are two pioneering works. SESS applies consistency regularization between teacher and student models with asymmetric augmentations, while 3DIoUMatch selects pseudo-labels using classification confidence, objectness, and IoU prediction. The significant performance improvements of 3DIoUMatch propelled the pseudo-label-based teacher-student frameworks to the forefront of SS3DOD \cite{3dioumatch,proficient,dds3d,atf3d,detmatch,hssda,ted_hssda,reliable,patchpillar,ateacher, notevery}. Proficient Teacher \cite{proficient} generates pseudo-labels by clustering bounding boxes from spatially and temporally augmented views. DDS3D \cite{dds3d} improved the recall rate through the dense pseudo-label generation while gradually decreasing the predefined threshold during training. ATF-3D \cite{atf3d} introduced a threshold searching mechanism that sets score thresholds by distance bins and classes, using predefined ratios of negative and positive samples. However, this approach discretizes distance and lacks consideration of the dynamic learning state. DetMatch \cite{detmatch} designed a consistency cost between 2D and 3D predictions and applied a manual threshold to filter pseudo-labels. Reliable Student \cite{reliable} proposed a robust learning method with class-aware target assignment and reliability-based loss softening, along with manual class-specific thresholds for pseudo-labels. A-Teacher \cite{ateacher} refined pseudo-labels gathering information from adjacent frames, while PTPM \cite{patchpillar} improved the teacher performance by dividing scenes into patches. These methods, however, still rely on handcrafted pseudo-label selection.
Recently, CSOT \cite{csot} developed a specialized model that synthesizes scenes by copying-pasting labeled objects to unlabeled scenes, demonstrating impressive performance. 
Note this technique is orthogonal to pseudo-labeling approaches. 
HSSDA \cite{hssda} is the state-of-the-art pseudo-labeling method that clusters three different scores of teacher predictions exceeding an IoU threshold with labels, generating two thresholds per score for hierarchical supervision.
This study aims to enhance the pseudo-label selection by learning the thresholding mechanism in SS3DOD, referring to HSSDA as the baseline.
\section{Methodology}
\label{sec:Method}

\begin{figure*}[t!] 
\centering
  \includegraphics[width=\textwidth]{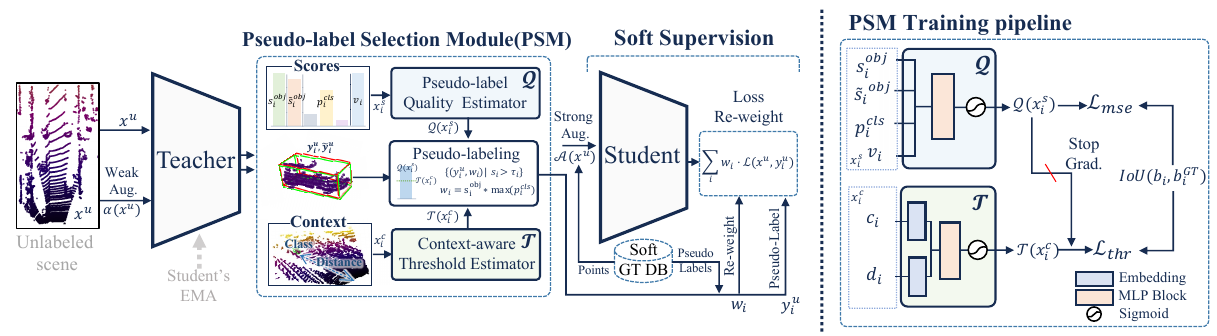}
\caption{Overview of the proposed framework, consisting of two main components: the Pseudo-label Selection Module (PSM), which selects pseudo-labels using the detector’s outputs and contexts, and Soft Supervision, which enhances robustness to pseudo-label noise. The PSM includes two neural networks, $\mathcal{Q}$ and $\mathcal{T}$, that predict pseudo-label quality and context-aware thresholds.}
\label{fig:main}
\vspace{-7.5pt}
\end{figure*}

\subsection{Teacher-student SSL Pipeline}
Semi-supervised 3D object detection involves training on a limited labeled dataset \( D^l \) and an abundant unlabeled dataset \( D^u \). The input point cloud consists of \( n \) points, and each point is characterized by 3D coordinates and additional information (e.g., color, intensity, and timestamps). The ground-truth annotation specifies objects in the labeled dataset using 7-dimensional parameters for 3D bounding boxes and a 1-dimensional object category.

While the 3D detector is trained with \( D^l \) in a supervised manner, the training process extends to semi-supervised learning (SSL) to incorporate \( D^u \). Mainstream SSL frameworks for 3D object detection involve four stages: (1) \textbf{Burn-in Stage}: Train the 3D detector on \( D^l \) to initialize both the teacher and student models. (2) \textbf{Pseudo-labeling Stage}: Generate pseudo-labels by filtering the teacher’s candidates on unlabeled data with weak augmentation $\alpha$. (3) \textbf{Semi-supervision Stage}: Compute the supervised and unsupervised losses from the student’s predictions on data with strong augmentation $\mathcal{A}$. (4) \textbf{Teacher Update Stage}: Update the teacher model using the EMA of the student model, 
\vspace{-0.1cm}
\begin{equation}
    \theta_t = \rho \cdot \theta_t+(1-\rho)\cdot\theta_s
\end{equation}
\(\theta_t\) represents the teacher parameters, which are updated based on the student parameters \(\theta_s\) using \(\rho\), the EMA momentum factor.

\subsection{Method Overview}
As illustrated in \cref{fig:main}, we introduce the Pseudo-Label Selection Module (PSM) within the teacher-student framework. The PSM reliably evaluates pseudo-label quality from various teacher outputs using the Pseudo-label Quality Estimator (PQE) and determines the threshold based on context-dependent score variations through the Context-aware Threshold Estimator (CTE). In the burn-in stage, the PSM is pre-trained using outputs from the teacher pipeline, which generates predictions for both original and weakly augmented scenes. From the teacher's outputs, we obtain instance-level predictions: objectness score \( s^{obj} \) and class distribution \( p^{cls} \) for the original scene, objectness score \( \tilde s^{obj} \) for the weakly augmented scene, and predicted bounding boxes \( b \) and \( \tilde b \) for original and weakly augmented scenes respectively. During the semi-supervision stage, the PSM is updated using the labeled data \(D^l\) to track the changes in the teacher’s state and perform adaptive pseudo-labeling.
 
Additionally, we introduce a supervision strategy called Soft Supervision that is robust to pseudo-label noises. This prevents bias to pseudo-label noises by re-weighting the loss with a joint confidence score. We generalize the hierarchical supervision \cite{hssda} that exploits dual-thresholds to a single threshold, and design Soft GT Sampling augmentation by modifying the GT sampling augmentation \cite{second}.

\subsection{Pseudo-label Selection Module (PSM)}
\label{sec:psm}
The proposed method aims to balance the quality and coverage of pseudo-labels using context-dependent multiple scores. 
If the ground truth (GT) labels are available, selecting pseudo-labels based on the Intersection over Union (IoU) with the GT bounding boxes is the most intuitive approach. GT-IoU provides a context-invariant measure of pseudo-label quality by indicating how close the pseudo-labels are to the actual ones. The PSM learns to predict or approximate GT-IoU-based pseudo-label selection using a labeled dataset \( D^l \) by two key components: the Pseudo-Label Quality Estimator (PQE) and the Context-Aware Threshold Estimator (CTE). 

{\bf Pseudo-label Quality Estimator (PQE).} Thresholding each score individually as in previous works \cite{3dioumatch, hssda} often misses high-quality pseudo-labels and reduces the diversity of labels. Instead of using individual scores, aggregating them into a single score accounts for the importance and combination of different score values. Filtering based on this fusion score helps increase the pseudo-label coverage while preserving their qualities.

PQE takes as input the feature vector $x^s_i = [s^{obj}_{i}, \tilde{s}^{obj}_{i}, p^{cls}_{i}, v_{i}]$, which consists of four components: the objectness score \(s^{obj}_i\), the auxiliary objectness score \(\tilde s^{obj}_i\), the classification probability \(p^{cls}_i\), and the IoU consistency \(v_i=IoU(b_i, \tilde b_i)\) for the $i$-th pseudo-label candidate. PQE, $\mathcal{Q}$, encodes this score feature to $\mathcal{Q}(x^s_i) \in [0,1]$, predicting the true quality of pseudo-labels, which is measured by GT-IoU i.e. $IoU(b_i, b_i^{GT})$, where \( b_i \) is the predicted pseudo-box and \( b_i^{GT} \) is the corresponding ground truth box.

The input data is passed onto a MLP module, yielding the predicted pseudo-label quality via a sigmoid function. The PQE is trained to minimize the mean squared error (MSE) loss between the GT-IoU and the predicted pseudo-label quality $\mathcal{Q}(x^s_i)$ over the pseudo-label candidates generated from the teacher before Non-Maximum Suppression (NMS). The training objective for PQE is as follows:

\begin{equation}
    \begin{aligned}
    \mathcal{L}_{PQE} = \frac{1}{N_l} \sum_i||\mathcal{Q}(x^s_i)-IoU(b_i,b_i^{GT})||^2_2
    \end{aligned}
\end{equation}
Where \( N_l \) is the number of pseudo-label candidates.

Inspired by the late candidate fusion networks in 3D Object Detection \cite{clocs, fastclocs}, PQE is designed to combine various scores of the teacher network and geometric associations at the output level. By aggregating diverse information, PQE provides a more reliable measure of pseudo-label quality. \cref{fig:scatter} shows that the PQE score exhibits a higher positive correlation with GT-IoU than other scores. By contrast, the classification confidence often undervalues high-quality pseudo-labels, increasing the risk of losing valuable samples. The reliability of the PQE score reduces the loss of valued samples during filtering and allows for wider coverage while maintaining quality.

{\bf Context-Aware Threshold Estimator (CTE).} 
While PQE provides a measure of pseudo-label quality, setting an appropriate threshold value for pseudo-label selection remains crucial. Since score distributions are context-dependent, the predicted pseudo-label quality $\mathcal{Q}(x^s_i)$ is also context-dependent. We consider the object class $c_i$ and distance $d_i$ as the context that influences the threshold, given the teacher’s current learning state $\theta_{t}$. The goal of Context-Aware Threshold Estimator (CTE) is to learn a context-aware threshold determination function $\mathcal{T}(c_i, d_i \mid \theta_{t})$ that mimics GT-IoU-based thresholding: 
\begin{equation}
    \label{eq:gtsim}
    \begin{aligned}
\mathcal{Q}(x^s_i)>\mathcal{T}(c_i,d_i|\theta_t) 
\quad \triangleq \quad
 IoU(b_i,b_i^{GT}) > \tau_{iou}
    \end{aligned}
\end{equation}
The threshold determination function $\mathcal{T}(\cdot)$ is implemented using a neural network. CTE takes the context inputs, represented as $x^c_i = [c_i, d_i]$, and includes an embedding layer for each context. It is followed by a MLP module and a sigmoid function, predicting the context-aware threshold. To train the CTE, we introduce a threshold error to evaluate the accuracy of the determined threshold, which serves as a loss function. The threshold error of the score $s$ and threshold $\tau$ is quantified as:
\vspace{-5pt}
\begin{equation}
    \resizebox{\columnwidth}{!}{$
    \begin{aligned}
    \mathcal{L}_{thr}(\tau,s,b,b^{GT}    
    ) = \begin{cases}
    ||\tau-s||^{2}_2 & \;\begin{bmatrix}(IoU(b,b^{GT}) \ge \tau_{iou} \land  s \le \tau) \ \lor \\ (IoU(b,b^{GT}) < \tau_{iou}  \land s > \tau) \end{bmatrix} \; \\
    0 & \;\text{otherwise}
    \end{cases}
    \end{aligned}
    $}
\end{equation}
We assign the L2 loss between the predicted pseudo-label quality \(s\) and the threshold \(\tau\) for the false cases in Eq.~(\ref{eq:gtsim}). Specifically, when a pseudo-label is correct (\(IoU(b,b^{GT})\ge\tau_{iou}\)) but \(\tau\) is higher than \(s\) (False Negative), a loss is applied. Conversely, when a pseudo-label is incorrect (\(IoU(b,b^{GT}) < \tau_{iou}\)) but \(\tau\) is lower than \(s\) (False Positive), it also contributes to the loss. A lower threshold error indicates a more optimal threshold in a global view, according to Eq. (\ref{eq:gtsim}). Through learning with the threshold errors of instances in a batch, the model progressively learns the threshold determination function $\mathcal{T}(\cdot)$. The training objective of the CTE is as follows:
\begin{equation}
\vspace{-8pt}
    \begin{aligned}
        \mathcal{L}_{CTE} = {1 \over N_l} \sum_i \mathcal{L}_{thr}(\mathcal{T}(x^c_i), \overline{\mathcal{Q}}(x^s_i), b_i, b_i^{GT})
    \end{aligned}
    \vspace{-5pt}
\end{equation} 
$\overline{\mathcal{Q}}(x^s_i)$ is the predicted pseudo-label quality that is stop-gradient, preventing gradient flow from the CTE to the PQE to avoid interference. Using $\mathcal{L}_{CTE}$, the model learns the appropriate context-specific threshold \(\mathcal{T}(x^c_i)\) for \(\mathcal{Q}(x^s_i)\).

\subsection{Soft Supervision}
Despite the proposed pseudo-labeling, unavoidable noises in pseudo-labels occur. To mitigate the impact of this noise, we propose a Soft Supervision that helps robust learning against pseudo-label noises. In the previous work HSSDA \cite{hssda}, the hierarchical supervision categorized pseudo-labels into a high-level and ambiguous-level. The loss for ambiguous-level pseudo-labels was softened, while high-level pseudo-labels were utilized for GT Sampling augmentation \cite{second}. This approach amplifies the influence of clean pseudo-labels and reduces the impact of noisy ones. Note that the pseudo-labels generated by PSM achieve a high precision and recall (see \cref{fig:quant}), making single-level pseudo-labels sufficient. We integrated and modified operations for both high-level (GT sampling augmentation) and ambiguous-level pseudo-labels (softened loss). Consequently, our supervision process is simplified yet reducing the effects of pseudo-label noises. 
The Soft Supervision includes Soft GT Sampling and Loss re-weighting.

{\bf Soft GT Sampling augmentation.} GT Sampling augmentation counteracts foreground sparsity by sampling GT from the labeled dataset and placing it into different frames. However, directly applying GT Sampling augmentation to inaccurate pseudo-labels results in excessive supervision signals containing noises, increasing the risk of overfitting to the noise. Therefore, we sample both the GT and their joint confidence score  $w = s^{obj} * \max(p^{cls})$, as in HSSDA \cite{hssda}. The joint confidence score is then used for the loss re-weighting to reduce the influence of noise. During SSL, we accumulate pseudo-labels in the Soft GT Database.

{\bf Loss re-weighting.} We soften the impact of noisy pseudo-labels using their associated joint confidence score $w$. These pseudo-labels are sourced from scene-generated pseudo-labels and samples from the Soft GT Database. This ensures the student focuses more on high-confidence pseudo-labels than noisy ones.

Soft Supervision simplifies and generalizes the hierarchical supervision \cite{hssda}, effectively addressing pseudo-label noises while maintaining the benefits of high-precision and high-recall pseudo-labels generated by PSM.

\subsection{Training Strategy}
During the burn-in stage, both the teacher and student networks are initialized after training the detector. The PSM is then trained using the teacher network’s output. Since CTE takes PQE as input, PQE’s learning states influence CTE. Both networks are trained together with a single optimizer, where PQE converges first and then CTE. The gradient of PSM does not backpropagate to the teacher network to avoid interfering with the detector training. 
In the semi-supervision stage, the student network is trained on both unlabeled and labeled datasets, while the PSM is trained exclusively on the labeled dataset. The total loss function incorporates three components: the labeled loss $\mathcal{L}_l$ and unlabeled loss $\mathcal{L}_u$ for the student network, along with $\mathcal{L}_{PSM}$ for the PSM network.
\begin{equation}
    \resizebox{\columnwidth}{!}{$
    \begin{aligned}
    \mathcal{L} =
    \underbrace{{1 \over N_l}\sum_i (
    \mathcal{L}^{cls}_i
    +\mathcal{L}^{reg}_i )}_{\mathcal{L}_l}
    +
    \underbrace{{1 \over N_u}
    \sum_i w_{i}(\mathcal{L}^{cls*}_i
    +\mathcal{L}^{reg*}_i )}_{\mathcal{L}_u}
    +\mathcal{L}_{PSM}
    \end{aligned}
    $}
\end{equation}
Where \(\mathcal{L}^{cls}\) and \(\mathcal{L}^{reg}\) represent the classification and regression losses using ground truth labels, respectively. For the unlabeled data, \(\mathcal{L}^{cls*}\) and \(\mathcal{L}^{reg*}\) are measured using the pseudo-labels by PSM and \(w_i\) serves as the joint confidence score for Soft Supervision. 
The PSM’s training loss $\mathcal{L}_{PSM}$ includes both PQE and CTE losses, as detailed in \cref{sec:psm}, where PSM is trained using teacher's pseudo-label candidates and ground truth labels for labeled scenes.
\begin{equation}
    \begin{aligned}
    \mathcal{L}_{PSM} = \mathcal{L}_{PQE} + \mathcal{L}_{CTE}
    \end{aligned}
\end{equation}
By minimizing the PSM loss, PSM evolves along with the learning state $\theta_t$ through the joint training. Pseudo-labels selected by CTE are used to train the student network, and the teacher network is updated via EMA of the student. The teacher’s predictions are then used to train CTE and PQE. This process repeats during training, establishing interactions between the student and PSM.

\begin{figure}[]
\centering 
\includegraphics[width=0.475\textwidth]{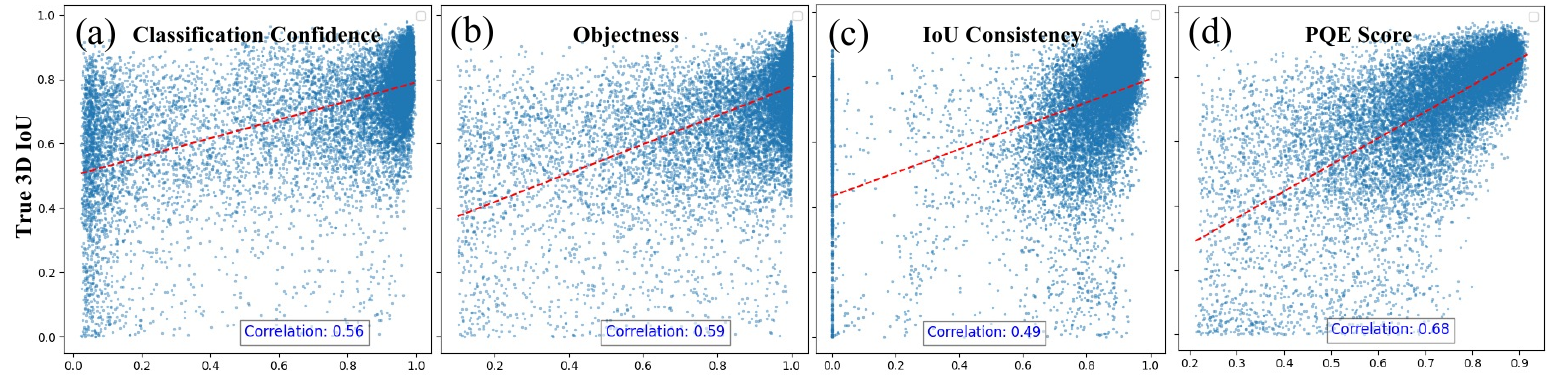}
\vspace{-0.5cm}
\caption{The correlation between GT-IoU and each score for KITTI 1\% split. (a) Classification confidence, (b) Objectness, (c) IoU consistency \cite{hssda}, and (d) the output score of PQE.}
\label{fig:scatter}
\vspace{-7.5pt}
\end{figure}

\section{Experiment}
\begin{table*}[htb!] 
    \centering
    \begin{tabular}{c|c|cccc|cccc} 
     \toprule
      \multirow{2}{*}{Model} & \multirow{2}{*}{Threshold} & \multicolumn{4}{c|}{1\%} & \multicolumn{4}{c}{2\%}  \\
      & & Car & Ped. & Cyc. & mAP & Car & Ped. & Cyc. & mAP  \\
     \hline
     \hline
     PV-RCNN (in \cite{3dioumatch}) & Detector & 73.5 & 28.7 & 28.4 & 43.5 & 76.6 & 40.8 & 45.5 & 54.3   \\ 
     \hline 
     3DIoUMatch \cite{3dioumatch} & Manual & 76.0 & 31.7 & 36.4 & 48.0 & 78.7 & 48.2 & 56.2 & 61.0  \\ 
    
     DDS3D \cite{dds3d}  & Manual & 76.0 & 34.8 & 38.5 & 49.8 & 78.9 & 49.4 & 53.9 & 60.7 \\
     Reliable Student \cite{reliable}  & Manual &  77.0 & 41.9 & 35.4 & 51.4 & 79.5 & 53.0 & 59.0 & 63.8 \\
     DetMatch \cite{detmatch} & Manual & 77.5 & \bf 57.3 & 42.3 & 59.0 & 78.2 & 54.1 & 64.7 & 65.6  \\
     \hline
     HSSDA \cite{hssda} & Auto &  80.9 & 51.9 & 45.7 & 59.5 & 81.9 & \bf 58.2 & 65.8 & 68.6 \\
    
     \bf Ours & Auto & \bf 81.3 & 47.0 & \bf 62.9 &  \bf 63.7 & \bf 82.0 & 56.8 & \bf 69.0 & \bf 69.3 \\
     \bottomrule
    \end{tabular} 
    \vspace{-7pt}
    \caption{Performance comparison on KITTI val set by PV-RCNN. All compared methods use PV-RCNN as the base detector. 
    The top row shows the result of the detector trained on the labeled-only dataset.}
    \label{tab:kitti_pv}
\vspace{-6pt}
\end{table*}

\begin{table*}[htb!]
\centering
\resizebox{\textwidth}{!}{
\begin{tabular}{c|c|cc|cc|cc|cc|cc|cc}
\toprule
\multirow{2}{*}{\begin{tabular}[c]{@{}c@{}}1\%\end{tabular}} &Threshold & \multicolumn{2}{c|}{Veh. (L1)} & \multicolumn{2}{c|}{Veh. (L2)} & \multicolumn{2}{c|}{Ped. (L1)} & \multicolumn{2}{c|}{Ped. (L2)} & \multicolumn{2}{c|}{Cyc. (L1)} & \multicolumn{2}{c}{Cyc. (L2)}\\
& &  AP & APH & AP & APH & AP & APH & AP & APH & AP & APH & AP & APH   \\ 
\hline \hline
PV-RCNN (in \cite{hssda})& Detector & 48.5 & 46.2 & 45.5 & 43.3 & 30.1 & 15.7 & 27.3 & 15.9 & 4.5 & 3.0 & 4.3&  2.9 \\
Voxel-RCNN (in \cite{hssda}) & Detector & 49.0 & 48.0 & 42.4 & 41.5 & 41.2 & 32.8 & 34.7 & 27.7 & 5.8 & 5.6 & 5.6 & 5.4 \\ 
\hline
DetMatch \cite{detmatch} \small{(by PV-RCNN)} & Manual & 52.2 & 51.1 & 48.1 & 47.2 & 39.5 & 18.9 & 35.8 & 17.1 & - & - & - & -  \\
*A-Teacher \cite{ateacher} \small{(by PV-RCNN)} & Manual & 56.5 & 54.5 & 49.2 & 47.5 & \bf 48.1 & \underline{27.3} & \bf 40.8 &  \underline{23.1} & \underline{35.1} & \bf 27.1 & \underline{33.7} & \underline{26.1}  \\
PTPM \cite{patchpillar} \small{(by PV-RCNN)} & Manual &\bf 61.5 & \bf 59.8 & \bf 53.7 & \bf 52.2 & 43.1 & 22.3 & 36.3 & 18.8 & \bf 35.7 & 17.9 & \bf 35.7 & \bf 34.3  \\

\hline
HSSDA \cite{hssda} \small{(by PV-RCNN)} & Auto & 56.4 & 53.8 & 49.7 & 47.3 & 40.1 & 20.9 & 33.5 & 17.5 & 29.1 & 20.9 & 27.9 & 20.0 \\
HSSDA \cite{hssda} \small{(by Voxel-RCNN)} & Auto & 54.9 & 54.1 & 48.3 & 47.5 & \underline{43.9} &\bf 37.8 & \underline{36.6} & \bf 31.6 & 17.5 & 16.7 & 16.7 & 16.0 \\

\bf Ours \small{(by Voxel-RCNN)} & Auto  & \underline{58.8} & \underline{57.3} & \underline{51.1} & \underline{49.8} & 30.6 & 16.5 & 25.5 & 13.8 & 34.8 & \underline{22.3} & 33.5 & 21.4  \\ \bottomrule
\end{tabular}}
\vspace{-7pt}
\caption{Experimental results on the Waymo validation set. * uses additional video information.
}
\label{tab:waymo_main}
\vspace{-12pt}
\end{table*}

\begin{table}[htb!]
\centering
\resizebox{1.0\columnwidth}{!}{
\begin{tabular}{ c|cccc|cccc} 
 \toprule
  \multirow{2}{*}{Model}  & \multicolumn{4}{c|}{1\%} & \multicolumn{4}{c}{2\%}  \\
  & Car & Ped. & Cyc. & mAP & Car & Ped. & Cyc. & mAP  \\
  \hline
 \hline
  Voxel-RCNN (in \cite{hssda})  & 74.0 & 19.0 & 37.0 & 43.3 & 76.5 & 40.2 & 39.9 & 52.2\\ \hline 

  HSSDA \cite{hssda} & \bf 81.7 & 43.9 & 48.3 & 58.0 & \bf 82.0 &  58.3 & 65.7 & 68.7  \\ 
\bf Ours & 81.4 & \bf 52.2 & \bf 61.5 & \bf 65.0 &  81.8 &  \bf 58.6 & \bf 70.6 & \bf 70.3   \\
 \bottomrule
\end{tabular}
}
\vspace{-8pt}
\caption{Experimental results on KITTI val set using Voxel-RCNN as the base detector for all methods.}
\label{tab:kitti_v}
\vspace{-21pt}
\end{table}

\subsection{Dataset and Evaluation Metric}
{\bf KITTI.}  
To evaluate the proposed framework, we utilize the KITTI 3D object detection benchmark \cite{kitti}, comprising 3,712 training scenes and 3,769 validation scenes. Following prior works \cite{hssda,detmatch}, we randomly select 1\% and 2\% of the full labeled data for the semi-supervised setting. For each ratio, we sample three distinct labeled sets and average the results across these sets to measure generalized performance independent of specific labeled sets. We evaluate three classes: Car, Pedestrian, and Cyclist—using Average Precision (AP) at 40 recall positions, applying Intersection over Union (IoU) thresholds of 0.7, 0.5, and 0.5 for each class, respectively.

{\bf Waymo.}  
We additionally evaluate our framework on the Waymo Open Dataset \cite{waymo}, which is the largest autonomous driving dataset containing 1,000 sequences. It includes 798 training sequences with approximately 150K point cloud samples and 202 validation sequences with about 40K samples. We sample 1\% of the training sequences (approximately 1.4K frames) for the semi-supervised setting. Due to the large scale of the Waymo dataset, we evaluate the results using a single split instead of averaging over three splits. We present AP and APH results at LEVEL 1 and LEVEL 2 difficulties for Vehicle, Pedestrian, and Cyclist classes.

\subsection{Implementation Details}
{\bf Network Architecture.} In PSM, CTE and PQE are lightweight 4-layer MLPs with channel dimensions $D_{MLP}=[16,32,32,1]$. For PQE, the score inputs are concatenated and then fed into the MLP. For CTE, the classes are linearly embedded to $D_{class}=8$, and distances are embedded to $D_{distance}=8$ dimensions using Fourier embedding \cite{fourieremb} after normalization. The embedded contexts are concatenated and then passed into the MLP.

{\bf Training Details.} Following prior works \cite{hssda}, we adopt PV-RCNN \cite{pvrcnn} and Voxel-RCNN \cite{voxelrcnn} as our baseline 3D detectors. During the semi-supervision stage, PSM and detector are jointly optimized using a single ADAM optimizer. The PSM is trained for 60 epochs with batch size 16, and we set the GT-IoU threshold \(\tau_{iou}=0.8\). See \cref{tab:abl_gt_threshold} for the effect of using different values.
We apply weak augmentation $\alpha$ with fixed transformations such as scaling, rotation, and flipping. For strong augmentation $\mathcal{A}$, we utilize stochastic transformations and Shuffle Data Augmentation \cite{hssda}.

\subsection{Main Results}
{\bf KITTI.}
We compare our method with state-of-the-art methods on the KITTI val set. \cref{tab:kitti_pv} presents the results based on the PV-RCNN. Compared to previous methods using PV-RCNN, our approach achieves the highest mAP, with absolute improvements of 20.2 and 15.0 at 1\% and 2\%, respectively. Notably, in the Cyclist class, we observe significant performance gains of 17.2 and 3.2 compared to the previous state-of-the-art at 1\% and 2\% settings. \cref{tab:kitti_v} shows the performance based on Voxel-RCNN. We observe similar behaviors of performance improvement. 
Under the setting of minimum labeled datasets 1\%, our method demonstrated substantial performance gains. Note also these results are obtained by the simpler pipeline that eliminates the dual-threshold based pseudo-label hierarchization and complex supervision strategies required by HSSDA \cite{hssda}. Moreover, learning PSM during SSL removes the need for iterative threshold recalculation.

{\bf Waymo.}

\cref{tab:waymo_main} presents the comparison results on the Waymo dataset. PTPM \cite{patchpillar} showed the best performance, followed by A-Teacher \cite{ateacher} and our method with comparable results. Note, however, PTPM \cite{patchpillar} and A-Teacher \cite{ateacher} use manual thresholds for pseudo-label selection, while our method significantly outperforms HSSDA, the other automatic threshold method, except on the Pedestrian class. PTPM \cite{patchpillar} mainly concerns designing an improved teacher network, and A-Teacher \cite{ateacher} refines pseudo-labels by incorporating additional information from adjacent frames rather than single images. These developments of the teacher network or the use of videos are orthogonal to the proposed idea.

The Pedestrian class exhibits particularly noisy patterns compared to other classes. The issues with the performance of the Pedestrian are known to the community. According to the official implementation of HSSDA \cite{hssda}, a different pseudo-label selection policy specific to Pedestrian is applied, whereas our method applies the same setting to all classes. Further discussions can be found in the supplementary.

\subsection{Ablation Studies and Analyses}

\begin{figure}[htb]
\centering 
\includegraphics[width=0.3\textwidth]{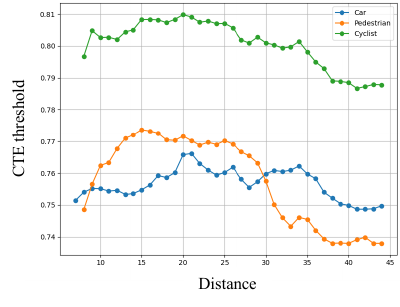}
\vspace{-13pt}
\caption{CTE thresholds by classes and distances.}
\label{fig:thrbydistmain}
\vspace{-10pt}
\end{figure}

In this section, we present experimental analyses to demonstrate the effect of our proposed framework. All results in this section are obtained using the KITTI 1\% split.

{\bf Contribution of Each Component.}
\cref{tab:abl_component} shows the results of each proposed component. Exp 1 presents the baseline results from HSSDA \cite{hssda}. Exp 2 demonstrates the effect of Pseudo-label Quality Estimation (PQE). The threshold generated by the Dual Threshold Generation \cite{hssda} for PQE is used for pseudo-label selection. We apply supervision without distinguishing between high-level and ambiguous-level pseudo-labels, and PQE alone outperforms the baseline. Exp 4 shows the effect of Context-aware Threshold Estimator (CTE). It demonstrates a significant performance gain with 4.2 mAP improvement over handcrafted thresholds. 

Exp 3 and Exp 5 illustrate the impact of Soft Supervision. They show meaningful performance improvements for the Pedestrian class, which has more noise in pseudo-labels compared to other classes. 

\begin{table}[t!]
\centering
\resizebox{0.9\columnwidth}{!}{
\begin{tabular}{c|cc|c|cccc}
\toprule
\multirow{2}{*}{Exp} & \multicolumn{2}{c|}{PSM} & \multirow{2}{*}{\shortstack{Soft \\ Supervision}} & \multirow{2}{*}{Car} & \multirow{2}{*}{Ped.} & \multirow{2}{*}{Cyc.} & \multirow{2}{*}{mAP} \\ 
 & PQE & CTE  &  &  &  &  &  \\ \hline
1 & - & - & - & 79.3 & 49.3 & 43.8 & 57.5 \\ 
2 & \checkmark & - & - & 80.6 & 43.8 & 59.4 & 61.2 \\
3 & \checkmark  & - & \checkmark & 81.1 & 47.0 & 60.3 & 62.8 \\
4 & \checkmark  & \checkmark & - & 81.3 & 50.9 & \bf 64.4 & \bf 65.5 \\
5 & \checkmark  & \checkmark & \checkmark & \bf 81.4 & \bf 52.2 & 61.5 & 65.0 \\ \bottomrule
\end{tabular}
}
\vspace{-8pt}
\caption{Ablation studies of each component on KITTI val set.}
\label{tab:abl_component}
\vspace{-8pt}
\end{table}

\begin{table}[t!]
\centering
\resizebox{0.9\columnwidth}{!}{
\begin{tabular}{c|cc|cccc}
\toprule
\multirow{2}{*}{Exp.} & \multicolumn{2}{c|}{Context}  & \multirow{2}{*}{Car} & \multirow{2}{*}{Ped.} & \multirow{2}{*}{Cyc.} & \multirow{2}{*}{mAP} \\ \cline{2-3}
 & Distance & Class &  &  &  &  \\ \hline
1 & - & \checkmark  & 80.8 & 59.3 & 67.7 & 69.3 \\ 
2 & \checkmark& - & \bf 82.0 & 50.4 & 68.5 & 67.0 \\
3 & \checkmark  & \checkmark &  81.8 & \bf 58.6 &  \bf 70.6 & \bf 70.3 \\ \bottomrule
\end{tabular}
}
\vspace{-8pt}
\caption{Ablation studies on contexts considered in CTE}
\label{tab:abl_context_main}
\vspace{-8pt}
\end{table}

\begin{table}[t!]
\centering
\begin{center}
\resizebox{0.85\columnwidth}{!}{
\begin{tabular}{ c|cccc } 
 \toprule
  GT-IoU Threshold & Car & Ped. & Cyc. & mAP \\
 \hline
 0.70 & 80.3 & 41.1 & 67.1 & 62.8 \\ 
  0.75 &  80.8 & 47.0 & \bf 67.4 & \bf 65.1 \\ 
 0.80 & \bf 81.4 & \bf 52.2 & 61.5 & 65.0 \\ 
 0.85 & 80.8 & 51.7 & 47.6 & 60.0 \\ 
 \bottomrule
\end{tabular}
}
\end{center}
\vspace{-8pt}
\caption{Effect of GT-IoU threshold $\tau_{iou}$}
\label{tab:abl_gt_threshold}
\vspace{-16pt}
\end{table}

{\bf Impact of Contexts in CTE.} We conducted an ablation study on different contexts in CTE. Using both class and distance contexts achieved the best accuracy, as in \cref{tab:abl_context_main}. We observed that using the distance improved the recall rate of pseudo-labels, which contributed to the performance gain for Car and Cyclist. \cref{fig:thrbydistmain} shows how the CTE thresholds vary across different classes and distances, similar to the score variations for distances as in \cref{fig:scoredist_distance} while exhibiting class-specific characteristics. Furthermore, unlike ATF-3D \cite{atf3d} and HSSDA \cite{hssda} where contexts are discretized, CTE operates in continuous context space, enabling a more flexible threshold determination mechanism without overfitting.

\begin{figure}[t!]
\centering 
\includegraphics[width=0.45\textwidth]{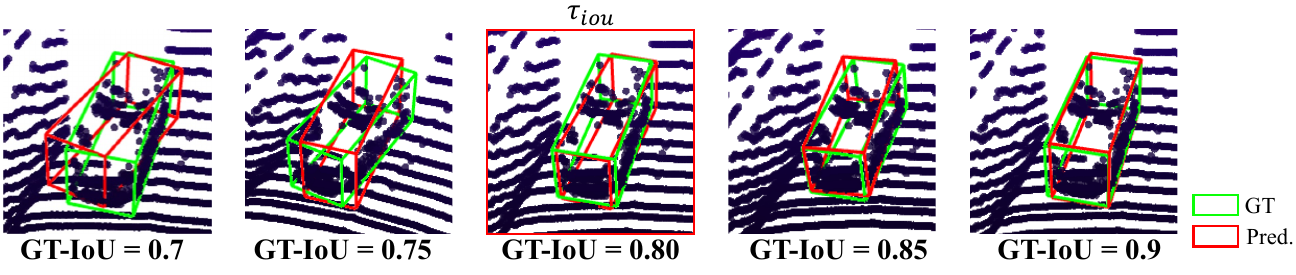}
\vspace{-0.3cm}
\caption{Visual comparison among different GT-IoU values.}
\label{fig:gtiou_com}
\vspace{-8pt}
\end{figure}

{\bf GT-IoU Threshold.}
We define the pseudo-labels among teacher predictions where the GT-IoU exceeds the threshold $\tau_{iou}=0.8$ for PSM training. While this is considered a hyperparameter, it is more general and interpretable than multiple score-level thresholds, which involve multiple dynamic scores ($s^{obj},\tilde s^{obj}, s^{cls},v$) and are sensitive and computationally complex. In contrast, the GT-IoU threshold is easier to set thanks to its geometrical and statistical intuitions, as shown in \cref{fig:gtiou_com}. The choice of this value as accurate labels is quite straightforward from visual overlaps and prior studies \cite{hssda} (see Supplementary for details). Existing automatic thresholding methods i.e. HSSDA, also involve a few hyperparameters to tune (e.g., the negative/positive sample ratios \cite{atf3d} and matching IoU threshold \cite{hssda}).
Given the value of $\tau_{iou}$, the CTE automatically and adaptively determines the score-level threshold while accounting for contextual factors. \cref{tab:abl_gt_threshold} shows the ablation study on the GT-IoU threshold $\tau_{iou}$. There is little performance change for the Car class with different values of $\tau_{iou}$. In contrast, the Pedestrian class exhibits a decline in performance as $\tau_{iou}$ decreases, while the Cyclist class performs poorly at higher $\tau_{iou}$ values. We set $\tau_{iou}$ as 0.8 which yields the most balanced performance among classes, and fixed for all datasets. Note the mAP is not sensitive to $\tau_{iou}$.

\begin{figure}[t!]
\centering 
\includegraphics[width=0.45\textwidth]{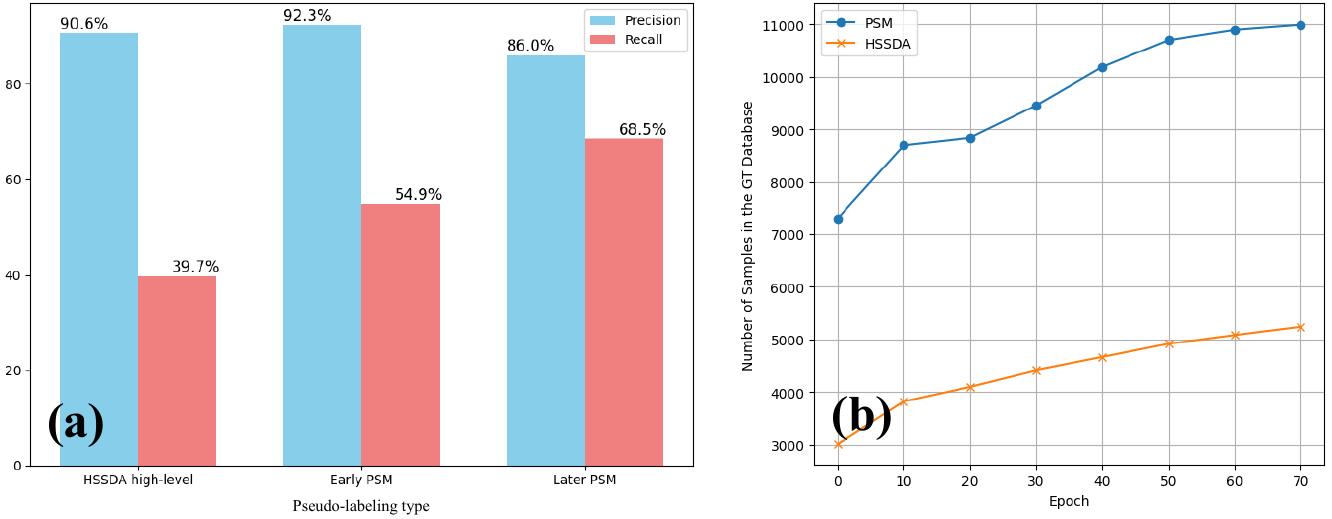}
\vspace{-0.3cm}
\caption{Quantitative comparisons of pseudo-label qualities on KITTI. PSM is pre-trained with the 1\% split.}
\label{fig:quant}
\vspace{-8pt}
\end{figure}

\begin{figure}[t!]
\centering 
\includegraphics[width=0.45\textwidth]{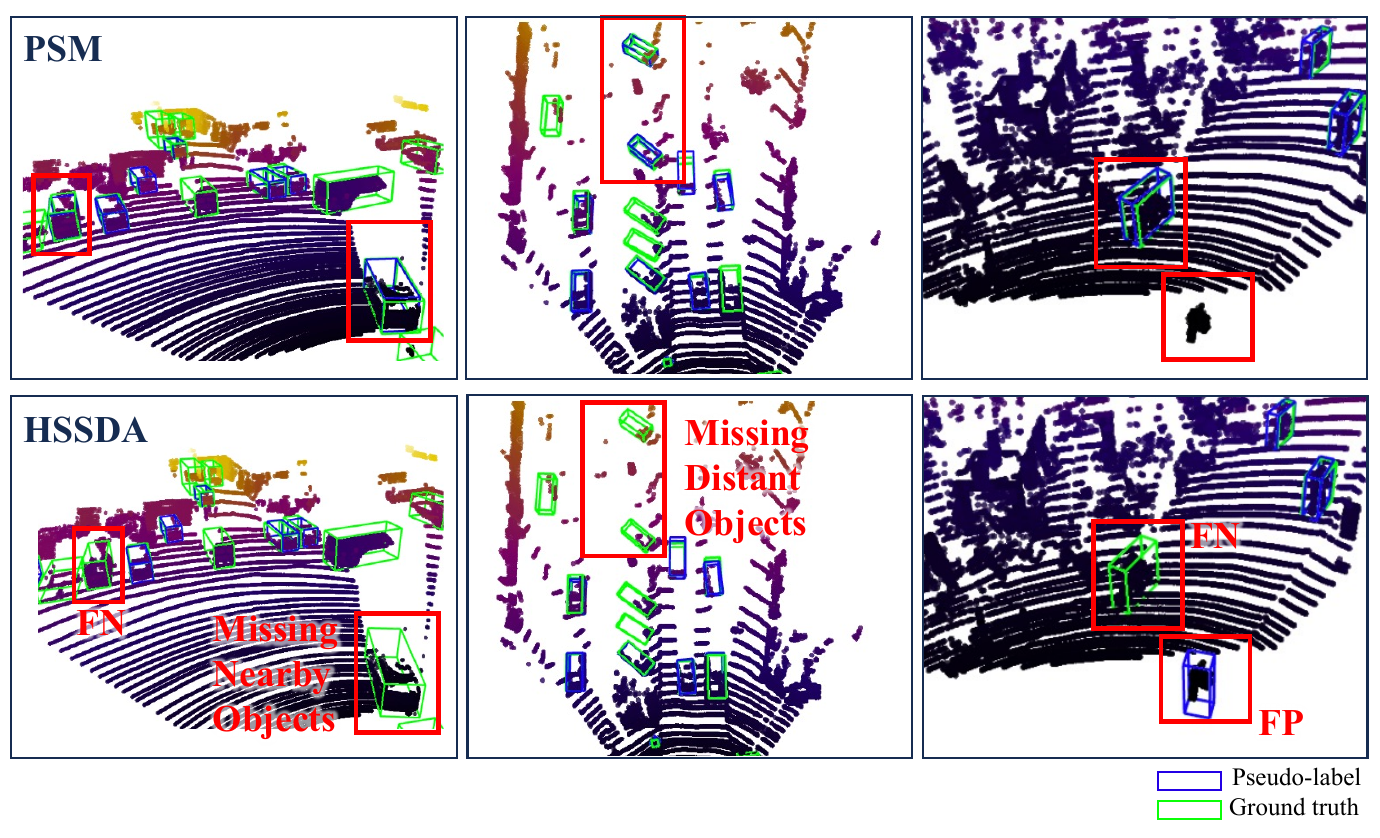}
\vspace{-0.3cm}
\caption{Qualitative comparisons of pseudo-labels on KITTI. PSM is pre-trained with the 1\% split.}
\label{fig:qual}
\vspace{-15pt}
\end{figure}

{\bf Quality of Pseudo-Labels.}
The quality and coverage of pseudo-labels can be quantified using precision and recall. As shown in \cref{fig:quant}\textcolor{RoyalBlue}{a}, PSM's pseudo-labels show 1.7 higher precision and 15.2 higher recall than HSSDA's high-level pseudo-labels. Furthermore, after 80 epochs of SSL, PSM's pseudo-labels show only a 6.3 decrease in precision while demonstrating a notable 13.6 increase in recall compared to HSSDA. Consequently, PSM selects more precise and diverse pseudo-labels through context-aware pseudo-labeling, as illustrated in \cref{fig:qual}. Our framework also stores a substantially larger number of pseudo-labels in the GT Database, as shown in \cref{fig:quant}\textcolor{RoyalBlue}{b}, providing the student with rich supervision signals.

\section{Conclusions}
In this paper, we propose a novel learning-based pseudo-labeling method that predicts pseudo-label quality and determines context-aware thresholds within the SSL framework. This approach enables the generation of a large volume of high-quality pseudo-labels. We also introduce Soft Supervision to prevent the student model from overfitting to pseudo-label noises. The extensive experiments and ablation studies support the effectiveness of our framework. In the future, we plan to extend the proposed pseudo-labeling to more complex SSL scenarios that involve richer pseudo-label contexts, such as multi-modal settings.

{\bf Acknowledgments.}
This work was supported by NST grant (CRC21011, MSIT), IITP grant (RS-2023-00228996, RS-2024-00459749, RS-2025-25443318, RS-2025-25441313, MSIT) and KOCCA grant (RS-2024-00442308, MCST).

{
    \small
    \bibliographystyle{ieeenat_fullname}
    \bibliography{main}
}

\clearpage
\setcounter{page}{1}
\setcounter{section}{0}
\setcounter{figure}{0}
\setcounter{table}{0}

\renewcommand{\thesection}{S.\arabic{section}}
\renewcommand{\thetable}{S\arabic{table}}
\renewcommand{\thefigure}{S\arabic{figure}}
\maketitlesupplementary

\begin{figure}[t!]
\centering 
\includegraphics[width=0.45\textwidth]{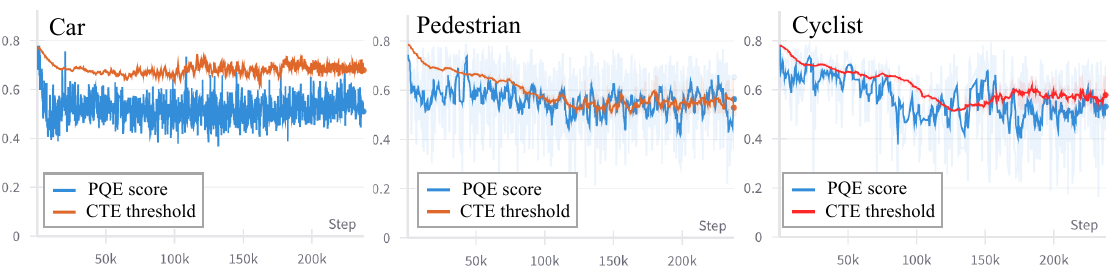}
\caption{CTE thresholds by classes and distances.}
\label{fig:learningstate}
\end{figure}

In the supplementary material, we provide additional details, visualizations, and analyses of the proposed framework. 

\section{Implementation Details}
\label{sec:s4}
We pre-trained Voxel-RCNN with a batch size of 16 for 400 epochs on KITTI during the burn-in stage, while for PV-RCNN we utilized pre-trained weights from HSSDA \cite{hssda}. The PSM based on Voxel-RCNN or PV-RCNN was pre-trained for 60 epochs with a batch size of 16 and 20, respectively. For the Waymo dataset, the Voxel-RCNN was pre-trained with a batch size of 4 for 50 epochs, and the PSM was pre-trained with a batch size of 8 for 12 epochs. During SSL, we used a batch size of 48 for 80 epochs on KITTI and a batch size of 16 for 10 epochs on Waymo. All SSL experiments were conducted using four 4090 GPUs. The PSM takes logits before the sigmoid and softmax as input during training. Other hyperparameters were adopted as specified in HSSDA \cite{hssda}.

\section{Analyses of PSM}

\hspace{1em}\textbf{Pseudo-label Quality Estimator (PQE).}
PV-RCNN \cite{pvrcnn} and Voxel-RCNN \cite{voxelrcnn} also incorporate a GT-IoU estimation module, similar to PQE. The key difference of PQE lies in that the pseudo-label quality is predicted more reliably by aggregating diverse information through a score fusion manner, including semantic scores and geometric consistency between original and augmented scenes. \cref{fig:scatter} in the main paper shows a stronger positive correlation between the PQE score and GT-IoU than the objectness score (i.e., predicted IoU). 

\begin{table}[t!]
\centering
\begin{center}
\begin{tabular}{ c|cc|c } 
 \toprule
  & True Positive & False Positive & Error rate\\
 \hline
  Car. & 9,307 & 1,240 & 12\% \\ 
  Ped. &  509 & 671  & \bf 57\%\\ 
  Cyc. &  144 & 33  & 19\% \\ 
 \bottomrule
\end{tabular}
\end{center}
\vspace{-0.1cm}
\caption{Error rates of different classes for high-confident predictions}
\label{tab:errorate}
\end{table}

\textbf{Context-aware Threshold Estimator (CTE).} 
The CTE threshold demonstrates its contextual encoding capabilities. \cref{fig:thrbydistmain} shows how the thresholds vary across different classes and distances. This enables PSM to effectively capture objects at both near and far distances (see \cref{fig:addqual}), thereby improving the recall rate (see \cref{fig:quant}\textcolor{RoyalBlue}{a}). As shown in \cref{fig:learningstate}, CTE adaptively determines thresholds in response to PQE scores updated during training, which reflects the teacher's evolving learning state.

\section{Performance on Pedestrian class}
\label{sec:s3}

\begin{figure}[t!]
\centering
  \includegraphics[width=0.45\textwidth]{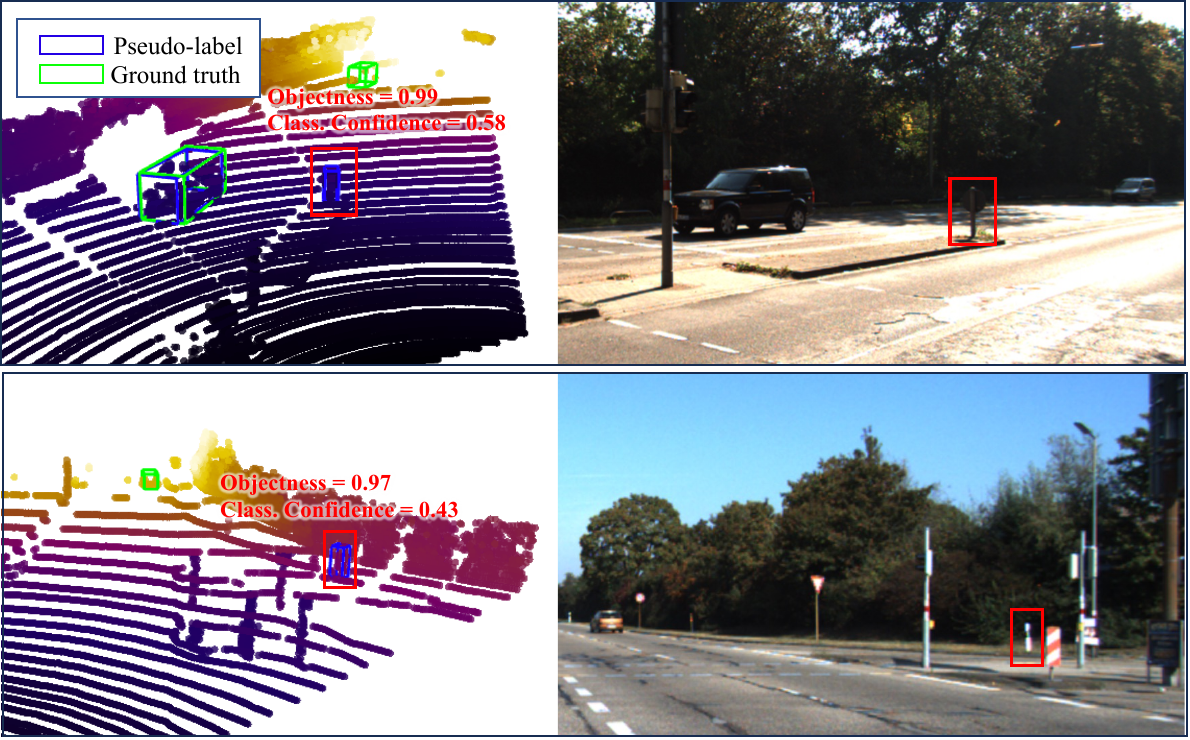}
  \caption{Failure cases of PSM for the Pedestrian class}
  \vspace{-0.1cm}
  \label{fig:failure}
\end{figure}

As shown in \cref{tab:errorate}, the Pedestrian class exhibits a unique pattern where the majority of high-confidence predictions ($s^{obj}>0.8$) are misclassifications. This occurs due to the difficulty in distinguishing objects like poles and signs from pedestrians given point cloud representation (see \cref{fig:failure}). Such overconfidence adversely impacts PQE’s ability to predict the GT-IoU and induces confirmation bias in the student model. Previous studies \cite{3dioumatch,dds3d,detmatch,reliable} suppress the pedestrian confirmation bias by maintaining a ratio between the labeled and unlabeled data (e.g., 1:1). On the other hand, HSSDA \cite{hssda} randomly samples batches without distinguishing between the labeled and unlabeled data, causing performance degradation due to pedestrian overconfidence. Consequently, the implementation of HSSDA available by the authors excludes the ambiguous-level pseudo-labels for the pedestrian class during SSL. In contrast, our method applies the same settings to all classes without any changes specific to the pedestrian class. We achieve comparable performance for the Pedestrian class on KITTI under the unified setting. 

\begin{figure}[t!]
\centering 
\includegraphics[width=0.45\textwidth]{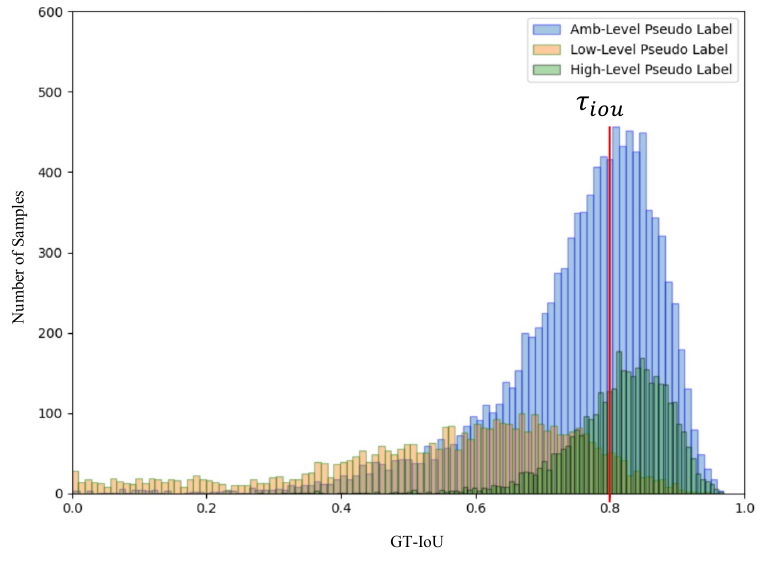}
\vspace{-0.3cm}
\caption{GT-IoU distribution of HSSDA's pseudo-labels.}
\label{fig:gtioudist}
\end{figure}

\section{GT-IoU Threshold}
\label{sec:s2}
We argue that our GT-IoU threshold ($\tau_{iou}$) is more interpretable and therefore easier to set than other score-based thresholds. As shown in \cref{fig:gtiou_com}, the quality of pseudo-labels can be intuitively assessed through the visual overlap between the pseudo-labels and GT. Additionally, \cref{fig:gtioudist} shows the GT-IoU distribution of pseudo-labels across the different levels in HSSDA. The distribution peaks at around 0.8, offering an intuitive basis for setting $\tau_{iou}$. Thanks to the generic $\tau_{iou}$, complex contextual factors from the multiple scores are automatically accounted for, without the need for manual selections as in prior works.

\begin{figure*}[t]
\centering
  \includegraphics[width=0.8\textwidth]{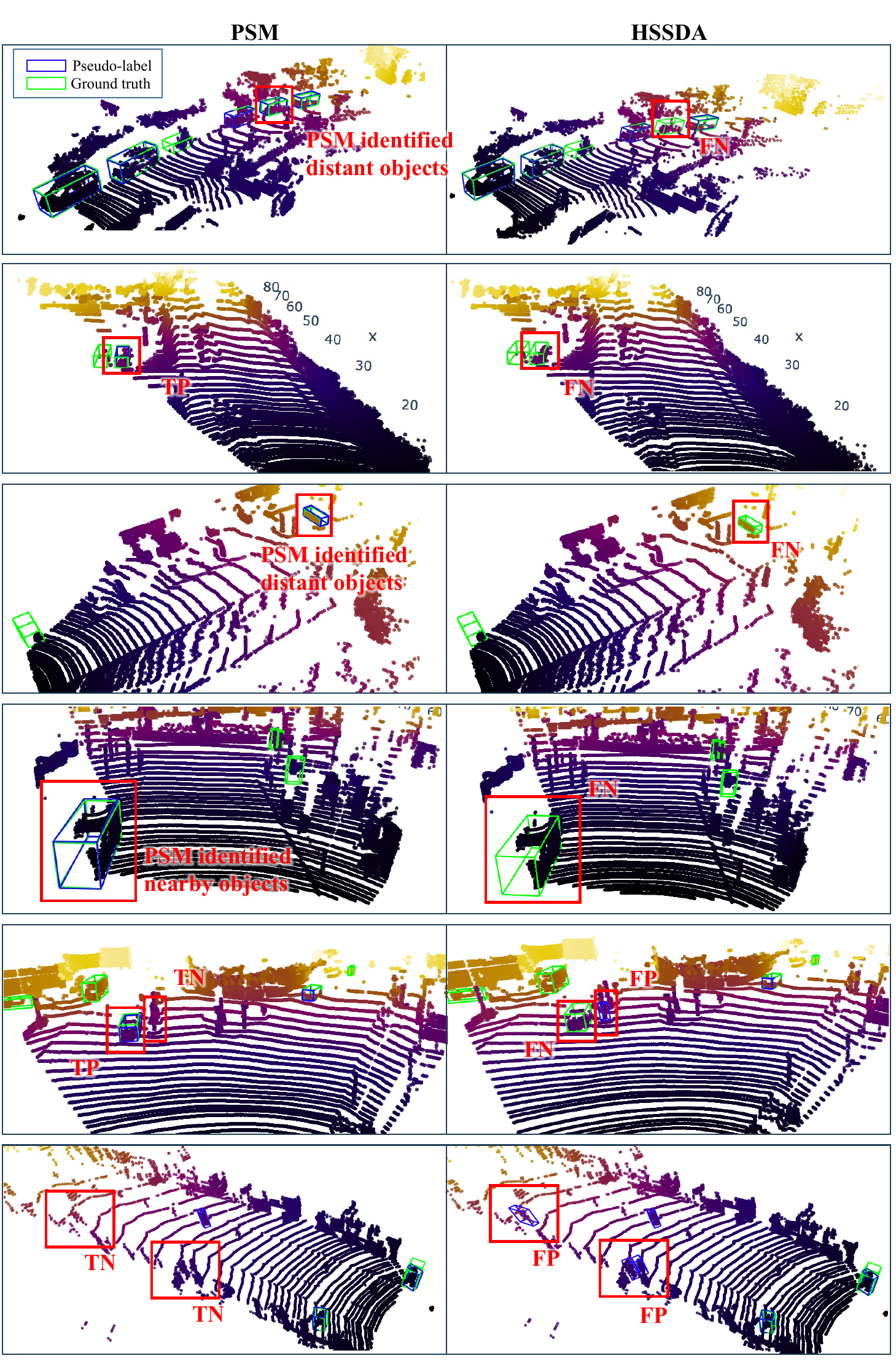}
  \caption{Additional qualitative results of pseudo-labels}
  \vspace{-0.1cm}
  \label{fig:addqual}
\end{figure*}

\section{Additional Qualitative Results}
\label{sec:s5}
\cref{fig:addqual} presents additional qualitative results of the pseudo-labels generated by PSM. We observe that PSM’s pseudo-labels maintain high quality while demonstrating wider coverage. Through context-aware thresholding, PSM selects pseudo-labels across a broad range of contexts. Additionally, by utilizing the score fusion and geometric consistency information from different views, PSM generates high-quality pseudo-labels.

\end{document}